\documentclass[11pt]{article}
\usepackage[margin=1in]{geometry}
\usepackage[T1]{fontenc}
\usepackage[utf8]{inputenc}
\usepackage{times}
\usepackage{microtype}
\usepackage{amsmath,amssymb,amsthm}
\usepackage{graphicx,booktabs,multirow,longtable,tabularx,array}
\usepackage{caption}
\usepackage{enumitem}
\usepackage{placeins}
\usepackage[numbers,sort&compress]{natbib}
\usepackage[hidelinks]{hyperref}
\usepackage{url}
\usepackage{tikz}
\usetikzlibrary{arrows.meta,positioning,fit}
\newcommand{\method}{DeciForm}

\title{DeciForm: Incorporating Task Structure into Decision Models}
\author{%
\begin{tabular}{c}
\textbf{Qiuyu Tian}$^{1,2}$ \quad
\textbf{Haojie Yin}$^{3}$ \quad
\textbf{Yingce Xia}$^{2}$ \quad
\textbf{Youyong Kong}$^{1}$ \\
\textbf{Zequn Liu}$^{2}$ \\
\small $^{1}$Southeast University, Nanjing, China \\
\small $^{2}$Beijing Zhongguancun Academy, Beijing, China \\
\small $^{3}$Duke Kunshan University, Kunshan, China
\end{tabular}%
}
\date{}
\hypersetup{pdftitle={DeciForm: Incorporating Task Structure into Decision Models},
  pdfauthor={Qiuyu Tian, Haojie Yin, Yingce Xia, Youyong Kong, Zequn Liu}}
\begin{document}
\maketitle

\begin{abstract}
Decision models unify candidate selection, Boolean judgment, and numerical scoring within a single model. Their answer spaces nevertheless carry different structures: classification candidates form an unordered set, while numerical scores have an intrinsic order and distance. We introduce \method{}, which incorporates these distinctions into computation and supervision. Candidate-free queries and isolated candidate branches support consistent classification under reordering. A log--spherical objective trains candidate probabilities, while a decimal-prefix readout combines individual-value and interval-weighted cumulative supervision for numerical scoring. Shared adapters, task-private residuals, and output-channel gains support joint adaptation. On a shared 15,000-request test set, \method{} improves all three decision metrics on Qwen and ModernBERT; source-normalized errors support the Score gains across the data mixture. Canonicalized joint encoding also eliminates tested order flips, while isolation further improves prediction metrics. Separate objective studies and main-configuration ablations examine the classification and numerical components. After one epoch of MIND-small adaptation, \method{} improves multi-positive news ranking on both backbones. These results support matching encoding and supervision to answer-space structure within a unified decision interface. Code is available at \url{https://github.com/roytian1992/DeciForm}.
\end{abstract}

\section{Introduction}
\label{sec:introduction}
Search and agentic systems repeatedly make decisions over explicitly specified answers. A system may select a passage, decide whether a condition holds, or assign a numerical quality score. These operations differ in their answer spaces, yet all require interpreting text together with instructions. Decision models expose a common interface for these tasks: given context, a question or rubric, and valid answers, the model returns a distribution over the answers. The Jev interface and open implementations such as Kev and Laya make this formulation concrete~\cite{typesafe2026,kev2026,laya2026}. Compared with generating and parsing arbitrary text, a typed answer distribution directly supports selection and scoring.

A shared interface does not make all answer spaces interchangeable. In candidate selection, the displayed list usually represents an unordered set. Reordering that list should change the positions of the output probabilities, but not the semantic decision. In numerical scoring, values instead have an intrinsic order and distance: confusing adjacent scores and confusing opposite ends of a rubric are different errors. These two structures call for different computational and learning choices.

Existing decision interfaces leave room to exploit both properties more directly. Jointly encoding a displayed list lets candidate positions and neighboring candidates influence its representations. Training numerical values only as unrelated categories supplies a correct-answer signal, but does not explicitly account for the distance between legal scores. The difficulty is to respect these distinctions while retaining shared language understanding and a single model for multiple decision types.

We introduce \method{}, which incorporates answer-space structure into encoding and supervision. For classification, a candidate-free query and independently encoded candidate branches produce scores without cross-candidate information flow. Consistent positional origins, canonical reductions, and semantic tie-breaking support deterministic decisions under reordering. A log--spherical objective supplements cross-entropy with a term that distinguishes concentrated from dispersed probability on incorrect answers. For numerical scoring, a decimal-prefix distribution represents the legal values, while individual-value and interval-weighted cumulative losses learn their probabilities. Shared low-rank updates, output-channel gains, and task-private residuals adapt the model to the three decision types.

The evaluation separates predictive quality from order consistency. A joint encoder given a deterministically sorted input can also eliminate external-order flips. We therefore compare isolation with both native-order and canonicalized joint encoding, and interpret the zero-flip audit separately from the accuracy differences. We also distinguish system-level comparisons, component removals in the main training configuration, and independent one-epoch studies of the classification and numerical objectives.

Across causal Qwen and bidirectional ModernBERT backbones, \method{} improves the reported metrics for candidate selection, Boolean judgment, and numerical scoring. Source-level normalized errors show that the Score improvement is not confined to a single large-range source. One-epoch adaptation to MIND-small further tests multi-positive ranking with complete candidate sets at evaluation. Together, these experiments support answer-space structure as a useful design principle for typed decision models.

Our contributions are threefold:
\begin{itemize}[leftmargin=*,itemsep=3pt,topsep=4pt]
\item An encoding design that isolates candidate computation, with matched canonicalized-joint controls that separate predictive benefits from deterministic ordering.
\item Task-specific probability objectives for unordered selection and distance-aware numerical scoring, including a multi-positive extension for ranking.
\item An evaluation across two backbones that combines mixed-task results, source-level errors, component and objective ablations, order audits, personalized news ranking, and computational cost.
\end{itemize}

\section{Related Work}
\paragraph{Ranking and candidate-order structure.}
Text rerankers can score query--passage pairs with cross-encoders or sequence-to-sequence relevance labels~\cite{nogueira2019bert,nogueira2020monot5}. Listwise methods model a candidate list~\cite{cao2007listnet}, and LLM rankers can produce permutations or score graded relevance labels~\cite{sun2023rankgpt,zhuang2024beyond}. These designs differ in how much candidates can interact and how an ordering is produced. Deep Sets and SetRank explicitly study set structure, including permutation-invariant functions and equivariant cross-document interactions~\cite{zaheer2017deepsets,pang2020setrank}. Our design uses isolated candidate representations and retains Boolean and numerical decisions in the same interface.

Candidate presentation also matters in multiple-choice language-model evaluation~\cite{pezeshkpour2024order}. AnyJev aggregates rotated readouts as part of its decision-model design~\cite{zhang2026anyjev}. We instead keep a candidate's context and positional origin consistent, prevent information exchange through the shared context, and canonicalize numerical execution. A canonically sorted joint encoder is an important control because stable input serialization can itself remove external-order dependence. Our evaluation distinguishes that stability property from prediction quality.

\paragraph{Typed decision interfaces and adaptation.}
Jev exposes a common interface for typed decisions, and Kev and Laya implement causal pointer and bidirectional-encoder systems~\cite{typesafe2026,kev2026,laya2026}. We build on this common setting to study the relation between the answer space, candidate computation, and training objective. Comparisons within each backbone retain shared records and candidate exposure while preserving the baselines' native main-task losses.

Low-rank adaptation changes model projections through compact learned updates~\cite{hu2022lora}. IA$^3$ uses learned activation scales, while DoRA separates weight magnitude from direction~\cite{liu2022ia3,liu2024dora}. LoRA+ assigns different learning rates to its two low-rank factors~\cite{hayou2024loraplus}. \method{} combines shared low-rank factors with output-channel gains and adds task-private residual capacity without reducing the shared rank. These are adaptation choices whose effects are tested by component removals.

\paragraph{Probability objectives for different answer spaces.}
Proper scoring rules define the population target of probabilistic prediction~\cite{gneiting2007proper}. Logarithmic and spherical scores provide different sensitivity to the shape of a categorical probability vector. For ordered outcomes, cumulative formulations use threshold events to reflect label order~\cite{mccullagh1980ordinal}. \method{} uses a log--spherical classification objective and a separate numerical objective over individual values and interval-weighted cumulative probabilities. Its prefix tree parameterizes the legal numerical support; the cumulative term supplies ordered supervision on that same distribution.

The objectives are evaluated through two complementary designs. Main-configuration removals connect them to the mixed-task system. Separate one-epoch Choice and Score studies compare losses and representations on task-specific mixtures, preserving the distinction between a categorical decision and numerical distance.

\section{DeciForm}
\label{sec:method}
\method{} uses a shared text backbone with computation and supervision matched to the answer type. Classification treats the candidate list as a presentation of an unordered set. Numerical scoring treats legal values as an ordered support with meaningful distances. The adaptation parameters are shared where useful and specialized by task type.

\begin{figure}[!htbp]
\centering
\includegraphics[width=\linewidth]{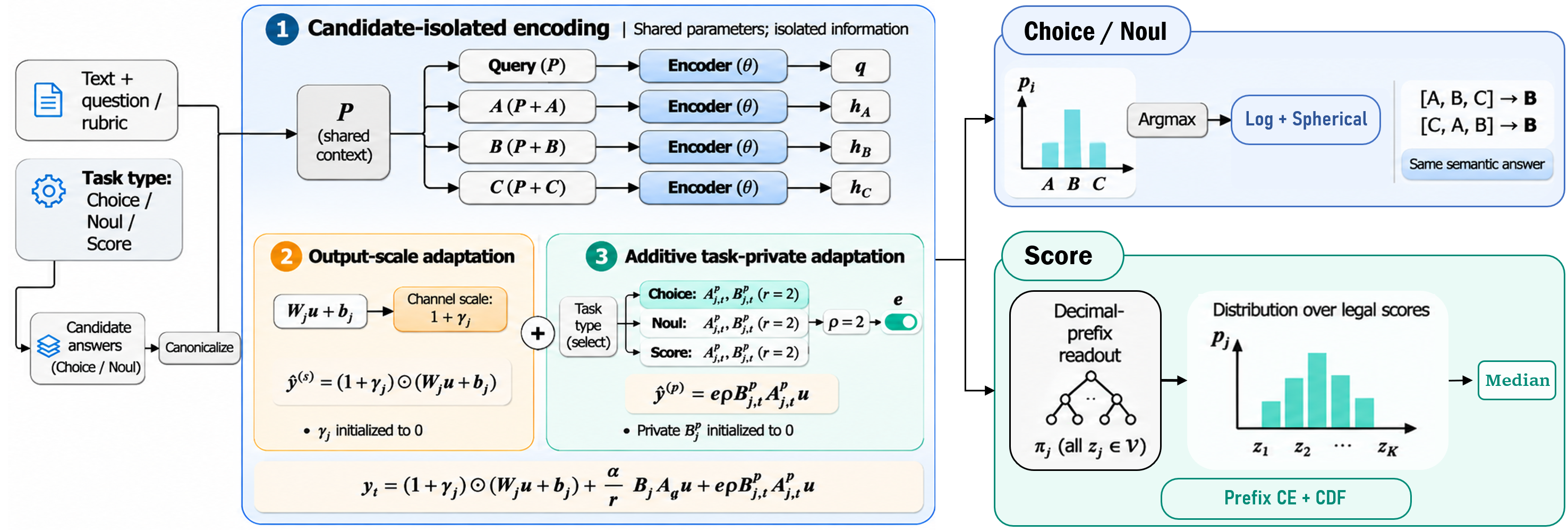}
\caption{Overview of \method{}. Query and candidate branches share backbone parameters while keeping their information separate. Output scaling and task-private residuals supplement the shared low-rank adapter. Choice/Noul use log--spherical supervision and argmax decoding; Score uses a decimal-prefix distribution, CE+CDF supervision, and median decoding. In the schematic, the query representation $q$, private rank $r=2$, and number of legal scores $K$ correspond to $h_\star$, $r_p=2$, and $L$ in the text, respectively.}
\label{fig:overview}
\end{figure}

Figure~\ref{fig:overview} summarizes the shared adaptation components and the distinct classification and numerical readouts.

\subsection{Task Interface and Notation}
Each request contains context text $x$, a question or rubric $q$, a task type $t$, and valid answers $C$. The model estimates
\begin{equation}
 p_\theta(y\mid x,q,t,C).
 \label{eq:interface}
\end{equation}
Choice selects a candidate, Noul makes a Boolean judgment between two answers, and Score predicts a legal numerical value. Classification uses $K=|C|$ candidates; Score uses sorted values $V=(v_1,\ldots,v_L)$. A numerical range is represented by a finite legal grid when necessary. Source names, benchmark identifiers, and example identifiers are not input features.

For classification, write $P=(x,q,t)$ for the candidate-free context. Although an implementation must serialize candidates in some order, the semantic request does not depend on that display order. For Score, the numerical ordering is part of the task definition. Table~\ref{tab:task-interface} summarizes these distinctions.

\begin{table}[!htbp]
\centering
\caption{The three decision types share a request interface but use different answer structure, supervision, and decoding. Multi-positive ranking extends the classification branch.}
\label{tab:task-interface}
\small
\begin{tabularx}{\linewidth}{@{}l*{3}{>{\raggedright\arraybackslash}X}@{}}
\toprule
Task & Answer structure & Training objective & Decision rule \\
\midrule
Choice & Unordered candidates & Log--spherical & Highest candidate probability \\
Noul & Two semantic answers & Log--spherical, $K=2$ & Highest answer probability \\
Score & Ordered numerical values & Value CE + weighted CDF & Posterior median \\
Multi-positive & Unordered candidates; several positives & Soft-target CE + normalized spherical term & Rank candidate probabilities \\
\bottomrule
\end{tabularx}
\end{table}

\subsection{Classification with Isolated Candidate Computation}
\label{sec:classification}
\paragraph{Candidate-free query and branch readouts.}
Joint encoding allows a candidate to depend on its neighbors and position in the displayed list. Instead, we compute a shared query from context alone and encode each candidate against that same context:
\begin{equation}
 h_\star=F_\theta(P,\texttt{query}),\qquad
 h_i=F_\theta(P,c_i).
 \label{eq:branch-encoding}
\end{equation}
Learned query and key projections produce a score for each candidate,
\begin{equation}
 s_i=\frac{Q(h_\star)^\top K(h_i)}{\sqrt d},\qquad
 p_i=\frac{\exp(s_i)}{\sum_{j=1}^{K}\exp(s_j)}.
 \label{eq:pointer}
\end{equation}
The projection dimension is $d$. Each score uses the same candidate-free query and only that candidate's branch. Softmax couples the final probabilities through normalization, but does not introduce cross-candidate information into the branch representations.

\paragraph{Causal and bidirectional implementations.}
Qwen branches reuse the same context state and positional origin. Every candidate therefore begins at the same offset relative to the shared context. ModernBERT requires an additional restriction because its context can otherwise attend to candidate tokens. At every layer, including local-attention layers, context tokens attend only to context; tokens in a candidate branch attend to context and their own branch. This blocks both direct candidate-to-candidate attention and indirect exchange through context tokens. The readout computes candidate scores separately, retaining the same separation after encoding.

\paragraph{Permutation behavior and deterministic inference.}
For a permutation matrix $\Pi$ acting on the displayed candidate order, the isolated construction gives
\begin{equation}
 s(P,\Pi C)=\Pi s(P,C),\qquad
 p(P,\Pi C)=\Pi p(P,C).
 \label{eq:equivariance}
\end{equation}
These statements describe equivariant scores and probabilities; the corresponding selected semantic answer is invariant when ties are resolved independently of display order. Appendix~\ref{app:math} gives the argument and the conditions required for deterministic execution.

Finite-precision computation requires care beyond the symbolic equations. Choice and Noul candidates are packed in UTF-8 byte order, reductions use a canonical order, and outputs are mapped back to the displayed order. Content-based tie-breaking removes dependence on the first displayed maximum. Score keeps its numerical order rather than applying lexical sorting.

Canonicalization and isolation address different parts of this construction. A deterministic joint encoder also receives the same internal input if external permutations are canonicalized first. Zero flips alone therefore cannot establish an advantage of isolation. The matched joint control in Section~\ref{sec:main-results} preserves canonicalization while changing the encoding, allowing predictive quality and runtime to be compared under the same order-stability outcome.

\subsection{Classification Objectives}
\label{sec:classification-loss}
\paragraph{Single-positive Choice and Noul.}
Cross-entropy supervises the probability of the correct candidate. We add a spherical-score term that also depends on the shape of the full probability vector:
\begin{equation}
 \mathcal L_{\rm cls}(p,y)
 =\frac{1}{1+\beta}\left[-\log p_y+
 \beta\sqrt K\left(1-\frac{p_y}{\|p\|_2}\right)\right],\qquad \beta\ge0.
 \label{eq:cls}
\end{equation}
At a fixed $p_y$, concentrating the remaining mass on fewer incorrect candidates increases $\|p\|_2$ and the spherical penalty. This gives a different training signal from cross-entropy alone. Logarithmic and spherical scores are proper scoring rules~\cite{gneiting2007proper}; their positive mixture remains proper for a fixed candidate set. This population property concerns the objective, rather than an empirical guarantee that a finite trained model is calibrated.

The $\sqrt K$ factor aligns the logit gradients of the spherical and logarithmic terms at a uniform prediction. The normalization by $1+\beta$ retains a common overall coefficient convention, and $\beta=0$ recovers cross-entropy. Appendix~\ref{app:math} provides the gradient calculation. Noul uses the same objective with $K=2$. Both tasks decode the candidate with maximum probability.

\paragraph{Multi-positive candidate ranking.}
For a training view containing positives $Y^+$, let $M=|Y^+|\ge1$ and define the normalized target
\begin{equation}
 \tau_i=\frac{\mathbb I[i\in Y^+]}{M},\qquad
 \mathcal L_{\rm softCE}=-\sum_i\tau_i\log p_i.
 \label{eq:multi-target}
\end{equation}
The ranking objective is
\begin{equation}
 \mathcal L_{\rm multi}
 =\frac{1}{1+\beta}\left[\mathcal L_{\rm softCE}
 +\beta\sqrt K\left(1-\frac{\tau^\top p}{\|\tau\|_2\|p\|_2}\right)\right].
 \label{eq:multi-loss}
\end{equation}
For a uniform target over positives, $\|\tau\|_2=1/\sqrt M$ and the normalized dot product is
\begin{equation}
 \frac{\tau^\top p}{\|\tau\|_2\|p\|_2}
 =\frac{\sum_{i\in Y^+}p_i}{\sqrt M\|p\|_2}.
\end{equation} When $M=1$, Equation~\eqref{eq:multi-loss} is exactly Equation~\eqref{eq:cls}. All positives contribute to one impression-level loss, rather than creating $M$ separately weighted training requests. For a fixed target distribution, both terms are minimized when $p=\tau$; the normalization gives the multi-positive term its stated scale. MIND adaptation uses this objective for \method{} and soft-target cross-entropy for the decision baselines.

\subsection{Distance-Aware Numerical Scoring}
\label{sec:objectives}
\paragraph{Individual values and cumulative thresholds.}
Numerical values carry more structure than category identities. A probability assigned far from the correct score crosses more numerical intervals than probability assigned to an adjacent score. We supervise both the value distribution and its cumulative probabilities so that training can use this structure.

Let $p_j$ and $\tau_j$ be predicted and target probabilities on increasing legal values $v_1<\cdots<v_L$. Define
\begin{equation}
 F_j=\sum_{i\le j}p_i,\qquad T_j=\sum_{i\le j}\tau_i,\qquad
 \omega_j=\frac{v_{j+1}-v_j}{v_L-v_1},\quad j<L.
 \label{eq:cdf}
\end{equation}
Here $F_j$ is the probability that the score is at most $v_j$, and $T_j$ is its target. The interval weights sum to one and assign larger weight to wider numerical gaps. The final threshold is excluded because $F_L=T_L=1$ carries no additional information.

The two losses and their combination are
\begin{align}
 \mathcal L_{\rm CE}&=-\sum_{j=1}^{L}\tau_j\log p_j,\label{eq:value-ce}\\
 \mathcal L_{\rm CDF}&=-\lambda\sum_{j=1}^{L-1}\omega_j
 \left[T_j\log F_j+(1-T_j)\log(1-F_j)\right],\label{eq:cdf-loss}\\
 \mathcal L_{\rm score}&=\frac{\mathcal L_{\rm CE}+\mathcal L_{\rm CDF}}{2}.
 \label{eq:score-loss}
\end{align}
The first term learns individual values; the second learns the ordered threshold events. Interval normalization makes their cumulative weighting depend on relative distances within the legal range. It does not change the native units used when measuring prediction errors.

\paragraph{Fractional supervision without rounding labels.}
If a label coincides with a legal value, the target is one-hot. If $y\in[v_a,v_{a+1}]$, only the two neighboring values receive mass:
\begin{equation}
 \tau_a=\frac{v_{a+1}-y}{v_{a+1}-v_a},\qquad
 \tau_{a+1}=\frac{y-v_a}{v_{a+1}-v_a},\qquad
 \tau_j=0\quad(j\notin\{a,a+1\}).
 \label{eq:interpolation}
\end{equation}
This interpolation preserves the original label as the expectation, $\sum_j\tau_jv_j=y$. Evaluation still compares the decoded prediction with the original label, including fractional values; it does not evaluate against a rounded training target.

\paragraph{A distribution over a decimal-prefix tree.}
We represent the legal values as leaves of a decimal-prefix tree. A shared prefix is computed once rather than separately for every complete value string. A single encoding of the context and legal support feeds the numerical readout. Frozen parameter rows associated with numeric tokens provide root logits, and a GRU models subsequent characters. At each node, probabilities are normalized over legal outgoing edges.

The probability of a value is the product along its path,
\begin{equation}
 p_j=\prod_{(a,c)\in E(v_j)}p(c\mid a,P,V),
 \label{eq:prefix}
\end{equation}
including the terminating-delimiter edge. Termination distinguishes a complete value from another legal value that extends its prefix. All legal values remain leaves, and the losses in Equations~\eqref{eq:value-ce}--\eqref{eq:cdf-loss} supervise the resulting value distribution. Appendix~\ref{app:math} explains its normalization.

\paragraph{Decoding for absolute error.}
Score returns the smallest legal value whose cumulative probability reaches one half:
\begin{equation}
 \widehat y=v_{j^\star},\qquad j^\star=\min\{j:F_j\ge1/2\}.
 \label{eq:median}
\end{equation}
This is a posterior median and minimizes expected absolute error under the predicted distribution. The distinction matters because the mode and mean optimize different decision criteria. The main probability-based Score comparisons use this same median rule.

\subsection{Shared and Task-Private Adaptation}
\label{sec:components}
\paragraph{Shared low-rank updates and output-channel gains.}
The task types share language understanding but need not use every projection identically. For projections receiving the same input $u$, we combine a low-rank update with learned output gains:
\begin{equation}
 \widetilde W_j(u)=\operatorname{Diag}(1+\gamma_j)(W_ju+b_j)
 +\frac{\alpha}{r}B_jA_gu.
 \label{eq:shared-adapter}
\end{equation}
$A_g$ is the input factor shared within a projection group, $B_j$ is projection-specific, and $\gamma_j$ gives one gain per output channel shared across task types. The gain acts on the original projection output, including its bias. When original weights are trainable, this scaling is a reparameterization rather than an enlargement of the representable linear function class.

\paragraph{Private residuals supplement the shared adapter.}
Each task type $t$ receives an additional rank-$r_p$ residual:
\begin{equation}
 \widehat W_{j,t}(u)=\widetilde W_j(u)+e\rho B^{\rm p}_{j,t}A^{\rm p}_{j,t}u,
 \qquad e\in\{0,1\}.
 \label{eq:private-adapter}
\end{equation}
The coefficient $\rho$ scales the private branch, and $e=0$ disables it. The full shared rank-$r$ adapter remains present; private capacity is not obtained by partitioning that rank. Shared parameters receive gradients from all task types, whereas a private branch receives gradients only from requests of its type. These pointwise projection changes preserve candidate isolation.

\paragraph{Request-level training.}
The appropriate objective is computed once per original request and then averaged over requests in a minibatch. This avoids weighting a request by the number of candidate branches it happens to contain. The fixed candidate views, parameter placement, initialization, and optimizer settings are specified in Section~\ref{sec:experiments} and Appendix~\ref{app:training}.

\section{Experiments}
\label{sec:experiments}
\subsection{Experimental Setup}
\textbf{Data and splits.}
We use 200,000 training requests, with shared development/test sets of
6,000/15,000. Development and test sets are task-balanced, giving 2,000 and 5,000 requests per type, respectively. Choice covers AG News, Amazon Reviews,
MNLI, and CFPB; Noul covers BoolQ, PAWS, QQP, HelpSteer3, and CommonsenseQA
2.0; Score covers HelpSteer, HelpSteer2, Prometheus feedback, SST-5, ASAP prompts
1--7, and
STS-B~\cite{zhang2015character,wang2018glue,helpsteer22024,prometheus2024,asap2012,cer-etal-2017-semeval}.
Related documents and rubric variants share split groups. Integer scores use
each rubric's full legal range; STS-B uses a 51-point grid on $[0,5]$ with
unrounded labels for MAE.

\textbf{Backbones and baselines.}
References are Kev for Qwen3.5-9B-Base and Laya for
ModernBERT-large~\cite{qwen35base,warner2025modernbert,kev2026,laya2026}.
Trained models share records and fixed candidate views, starting from pretrained
backbones and fresh decision parameters. Qwen trains adapters/readouts;
ModernBERT and Laya also tune the encoder. Main experiments retain native baseline losses
and both Laya layouts. We also test Direct answer (LoRA) and
three zero-shot readouts of post-trained Qwen3.5-9B~\cite{qwen35posttrained}.
For Score, probability models use posterior medians; generation uses numeric
JSON outputs.
Likelihood normalizes complete legal JSON-string probabilities across
answers~\cite{jiang2021know}; local normalization instead renormalizes valid
next-token continuations in an answer trie before scoring complete
paths~\cite{geng2023grammar,park2024grammaraligned}.

\textbf{Training and candidate exposure.}
All non-zero-shot methods train on the same 200,000 requests for three epochs
with one training seed per model. Both \method{} models use the full design in
Section~\ref{sec:method}, shared rank $12$ ($\alpha=24$), private rank $2$
per type, and 256-dimensional decision projections/prefix GRU. Private
residuals cover Qwen's last eight and all 28 ModernBERT layers. AdamW uses
global batch $64$ and peak rates $2\times10^{-5}$ for shared adapters,
pointers, and the ModernBERT encoder, and $10^{-3}$ for numeric readout/output
gains. All Qwen-based \method{} variants use LoRA+~\cite{hayou2024loraplus}
with $\eta_B/\eta_A=16$.

Choice training fixes one positive and at most three negatives, with identical
sampled candidates across methods; Noul/Score use complete supports.
Development/test evaluation retains all candidates and full input text.
Output scaling covers Qwen attention/DeltaNet outputs and MLP down projections,
and ModernBERT attention/MLP outputs. Gains and shared/private adapter output
factors start at zero. We set $\rho=2$, $e=1$, $\beta=0.75$, and
$\lambda\simeq3.04$.

\textbf{Metrics.}
Choice/Noul use accuracy. Score uses original labels to report native-unit MAE
over 5,000 requests and source-macro NMAE, $(1/12)\sum_s\mathrm{MAE}_s/(U_s-L_s)$,
where $[L_s,U_s]$ is the legal range and ASAP prompts are separate sources.
Reverse/Any count semantic prediction flips under reversal/any of reversal,
rotation, and shuffling. Noul has two orders; Score retains numerical order.
Reported 95\% CIs use 10,000 paired percentile cluster-bootstrap draws
(random seed 42), grouping main requests by shared document/prompt split group
and MIND impressions by user. Each draw retains the metric's original averaging;
intervals quantify test-sampling uncertainty conditional on the trained models.

The experimental settings answer different questions. Table~\ref{tab:study-map} distinguishes their training budgets and aggregation units; Appendix~\ref{app:training} consolidates the configuration details.
\begin{table}[!htbp]
\centering
\caption{Experimental settings. The independent objective studies use different mixtures from the main experiment; MIND begins from the mixed-task checkpoints.}
\label{tab:study-map}
\small
\begin{tabularx}{\linewidth}{@{}l*{3}{>{\raggedright\arraybackslash}X}@{}}
\toprule
Setting & Training & Evaluation & Main purpose \\
\midrule
Mixed-task main & 200,000 requests; 3 epochs & 15,000 requests; full support & Complete-system and component comparisons \\
Choice objective study & CFPB, Banking77, CLINC150; 1 epoch & Three-source macro accuracy & Compare classification losses \\
Score objective study & ASAP 1--8, STS-B, WMT; 1 epoch & Ten-source macro NMAE & Compare score representations and losses \\
MIND adaptation & 156,965 impressions; 1 epoch & 73,152 development impressions & Multi-positive personalized ranking \\
\bottomrule
\end{tabularx}
\end{table}

\subsection{Main Decision Results}
\label{sec:main-results}
\begin{table}[!htbp]
\centering
\caption{Main results. Fine-tuning: three epochs on 200,000 requests.
Score NMAE is source-macro; latency follows Section~\ref{sec:cost}.}
\label{tab:main}
\small
\setlength{\tabcolsep}{5pt}
\begin{tabular}{@{}llrrrr@{}}
\toprule
Setup & Model & Choice & Noul & Score & Score \\
 & & Acc (\%) $\uparrow$ & Acc (\%) $\uparrow$ & MAE $\downarrow$ & NMAE $\downarrow$ \\
\midrule
\multirow{3}{*}{\shortstack[l]{Qwen\\(fine-tuned)}} & Kev & 76.04 & 87.86 & 0.4860 & 0.0914 \\
 & Direct answer (LoRA) & 75.96 & 87.16 & 0.4980 & 0.0937 \\
 & \method{} & \textbf{77.38} & \textbf{88.92} & \textbf{0.4474} & \textbf{0.0850} \\
\midrule
\multirow{3}{*}{\shortstack[l]{ModernBERT\\(fine-tuned)}} & Laya sequential & 73.74 & 81.08 & 0.5866 & 0.1104 \\
 & Laya parallel & 73.52 & 80.84 & 0.6134 & 0.1154 \\
 & \method{} & \textbf{76.18} & \textbf{86.42} & \textbf{0.4814} & \textbf{0.0908} \\
\midrule
\multirow{3}{*}{\shortstack[l]{Instruct\\(zero-shot)}} & Direct answer & 68.04 & 77.42 & 0.9826 & 0.1848 \\
 & Likelihood & 67.36 & 78.34 & 0.9542 & 0.1795 \\
 & Local normalization & 67.46 & 78.28 & 0.9538 & 0.1794 \\
\bottomrule
\end{tabular}
\end{table}

\textbf{General decision benchmarks.}
\method{} improves all three tasks within each backbone (Table~\ref{tab:main}).
For Qwen versus Kev and ModernBERT versus Laya parallel, respectively,
Choice gain CIs are $[0.20,2.48]$ and $[1.48,3.84]$ percentage points;
Noul CIs are $[-0.12,2.24]$ and $[4.32,6.84]$.

For Score, native-MAE difference CIs are $[-0.0484,-0.0288]$ for Qwen and
$[-0.1506,-0.1134]$ for ModernBERT. Macro-NMAE CIs are
$[-0.0089,-0.0039]$ and $[-0.0308,-0.0184]$, respectively.
Score improves on 11/12 and 12/12 sources, respectively; Qwen's exception
is ASAP prompt 3. Appendix~\ref{app:source-score} gives every source count, legal range, MAE, and NMAE for all nine main-table models. These are system-level comparisons; zero-shot references
also differ in training.

ModernBERT's sequential Laya layout is stronger than its parallel layout in the main table, and \method{} improves over both. Relative to Laya sequential, Choice and Noul gains are 2.44 and 5.34 percentage points; relative to Laya parallel, the corresponding gains are 2.66 and 5.58 points. The reported paired intervals use Laya parallel, so their endpoints apply to that named comparison. For Qwen, the Noul point estimate improves, but its interval includes zero. This task-specific uncertainty remains visible alongside the positive Choice and Score comparisons.

The source-level Score results help distinguish range effects from broader improvement. Each ASAP prompt is treated as a separate source, and the normalized metric gives all twelve sources equal weight. Qwen's ASAP 3 MAE changes from 0.2850 to 0.2900, whereas its other eleven source errors decline. Both native and normalized aggregates favor \method{} on each backbone. The full uncertainty table and resampling definitions appear in Appendix~\ref{app:bootstrap}.

\begin{table}[!htbp]
\centering
\caption{Component and objective ablations on the main test set.}
\label{tab:ablation}
\small
\setlength{\tabcolsep}{5pt}
\begin{tabular}{@{}llrrr@{}}
\toprule
Backbone & Configuration & Choice & Noul & Score \\
 & & Acc (\%) $\uparrow$ & Acc (\%) $\uparrow$ & MAE $\downarrow$ \\
\midrule
\multirow{7}{*}{Qwen3.5-9B} & Full & \textbf{77.38} & \textbf{88.92} & \textbf{0.4474} \\
 & Joint encoding & 76.50 & 88.32 & 0.4542 \\
 & Joint + canonical & 76.72 & 88.48 & 0.4534 \\
 & w/o Output-scale & 77.14 & 88.72 & 0.4618 \\
 & w/o Task-private & 76.98 & 88.52 & 0.4558 \\
 & Classification CE & 76.70 & 88.56 & 0.4512 \\
 & Score no-CDF & 77.32 & 88.88 & 0.4724 \\
\midrule
\multirow{7}{*}{ModernBERT-large} & Full & \textbf{76.18} & \textbf{86.42} & \textbf{0.4814} \\
 & Joint encoding & 75.08 & 85.44 & 0.4918 \\
 & Joint + canonical & 75.34 & 85.68 & 0.4906 \\
 & w/o Output-scale & 75.88 & 85.96 & 0.5054 \\
 & w/o Task-private & 75.70 & 85.80 & 0.4938 \\
 & Classification CE & 75.42 & 85.94 & 0.4876 \\
 & Score no-CDF & 76.10 & 86.36 & 0.5168 \\
\bottomrule
\end{tabular}
\end{table}

\textbf{Component ablations.}
Joint controls retain the adapters, objectives, and three-epoch budget
(Table~\ref{tab:ablation}). Joint + canonical sorts Choice/Noul candidates by
UTF-8 bytes during training and inference, uses sequential positions, inverse
output mapping, and the same content-based tie-breaking; Score stays numerically
ordered. Isolation outperforms both joint variants. Removing private residuals
hurts classification more, while removing output scaling increases Score MAE more. These are removal-style comparisons within the same mixed-task training setup. Their point estimates describe the tested configuration; small differences are not treated as measurements of training-seed robustness.

\begin{table}[!htbp]
\centering
\caption{Prediction flip rates (\%) on the full-support test set.
Rev. compares reversed order; Any counts any tested reordering.
An em dash means unreported.}
\label{tab:order}
\small
\setlength{\tabcolsep}{5pt}
\begin{tabular}{@{}llrrr@{}}
\toprule
Backbone & Model & \multicolumn{2}{c}{Choice $\downarrow$} & Noul $\downarrow$ \\
 & & Rev. & Any & Rev./any \\
\midrule
\multirow{3}{*}{Qwen3.5-9B} & Kev & 6.84 & 14.24 & 4.36 \\
 & Direct answer (LoRA) & 10.46 & 16.64 & \textemdash \\
 & \method{} & \textbf{0.00} & \textbf{0.00} & \textbf{0.00} \\
\midrule
\multirow{3}{*}{ModernBERT-large} & Laya sequential & 8.12 & 13.26 & 10.48 \\
 & Laya parallel & 0.20 & 0.38 & 0.24 \\
 & \method{} & \textbf{0.00} & \textbf{0.00} & \textbf{0.00} \\
\midrule
\multirow{3}{*}{Instruct} & Direct answer & 21.24 & 41.28 & \textemdash \\
 & Likelihood & 21.96 & 42.62 & 2.46 \\
 & Local normalization & 21.62 & 41.98 & 2.44 \\
\bottomrule
\end{tabular}
\end{table}

\textbf{Candidate-order consistency.}
\method{} and canonicalized joint encoding both have zero classification flips
under tested reorderings. Native-order joint encoding has Choice-any rates of
13.38\% and 12.36\%, and Noul reversal rates of 4.08\% and 9.48\%,
for Qwen and ModernBERT, respectively.
Sorting therefore suffices for this stability audit; isolation additionally
improves prediction metrics. Deterministic sorting gives the joint encoder the
same internal input after external reordering, so zero flips alone cannot
distinguish its encoding from isolation. Table~\ref{tab:order} reports the other baselines.

\subsection{Personalized News Recommendation}
\label{sec:news-recommendation}
MIND-small evaluates ranking conditioned on a user's clicked news history~\cite{wu2020mind}. Each impression supplies a history, candidate articles, and click labels. It complements single-answer Choice with variable candidate sets and potentially multiple positives. We use the official 156,965 training and 73,152 development impressions, representing articles by title, abstract, and category.

Each impression has a fixed training view containing all clicked articles and at most three sampled unclicked articles. Uniform targets over the positives and one loss per impression give equal training weight to impressions. \method{} uses Equation~\eqref{eq:multi-loss} with $\beta=0.75$; Kev and Laya parallel use soft-target cross-entropy. All models adapt their respective mixed-task checkpoints for one epoch with a single seed, identical candidate views, and the same batch order. We evaluate the final checkpoint; development data are used only for evaluation.

Evaluation retains the complete user history and original candidate set. AUC, MRR, nDCG@5, and nDCG@10 are averaged over impressions. MRR averages reciprocal ranks across all clicked articles within each impression, following the official script~\cite{msnews2020evaluation}; Appendix~\ref{app:training} gives the formula. History TF-IDF and zero-shot Qwen likelihood provide reference readouts.

Table~\ref{tab:mind_results} shows improvements in all four metrics on both backbones. Qwen gains 5.53 MRR and 6.93 nDCG@5 percentage points over Kev. ModernBERT gains 1.03 and 1.26 points over Laya parallel, with a 95\% MRR-gain interval of $[0.78,1.28]$ points. The rank-sensitive metrics support improved placement of clicked articles near the top. These system comparisons reflect differences in both the mixed-task checkpoints and the adaptation objectives.

\begin{table}[!htbp]
\caption{MIND-small development results for personalized news recommendation.
All models use complete original candidate sets; metrics are percentages.}
\label{tab:mind_results}
\centering
\begin{tabular}{lcccc}
\toprule
Model & AUC $\uparrow$ & MRR $\uparrow$ & nDCG@5 $\uparrow$ & nDCG@10 $\uparrow$ \\
\midrule
History TF-IDF & 58.74 & 27.92 & 29.85 & 35.82 \\
Qwen likelihood (zero-shot) & 56.81 & 27.15 & 28.99 & 34.71 \\
\midrule
Laya parallel & 69.58 & 33.95 & 37.68 & 43.92 \\
\method{} (ModernBERT) & \textbf{69.91} & \textbf{34.98} & \textbf{38.94} & \textbf{45.10} \\
\midrule
Kev & 66.63 & 30.89 & 33.68 & 40.24 \\
\method{} (Qwen) & \textbf{72.34} & \textbf{36.42} & \textbf{40.61} & \textbf{46.85} \\
\bottomrule
\end{tabular}
\end{table}

\subsection{Separate Task-Specific Objective Studies}
\label{sec:objective-studies}
The main experiment trains all three decision types together. We also conduct separate one-epoch studies to examine classification and numerical objectives in their respective settings. These studies use different data mixtures from the main experiment. Within each study, \method{} fixes isolation, scaling, private adaptation, candidate views, data order, and schedules; reference systems receive matched training exposure. Their aggregate metrics are therefore interpreted within each study.

\subsubsection{Large-Candidate Choice}
The Choice study trains jointly on CFPB ($K=21$), Banking77 ($K=77$), and CLINC150 ($K=150$)~\cite{casanueva2020banking77,larson2019clinc}. Each fixed training view contains one positive and three negatives, whereas evaluation ranks the complete candidate set. This creates a distinction between the small candidate view used for each training request and the larger answer set used at inference.

We compare cross-entropy and log--spherical training within \method{} on both backbones. ModernBERT additionally compares the RLCD objective used by Laya. Kev and Laya parallel are reference systems, while changing only the loss within \method{} is the objective control. Table~\ref{tab:choice-objectives} reports CFPB and CLINC150 accuracy together with the equal-weight macro average over all three sources.

\begin{table}[!htbp]
\centering
\caption{One-epoch Choice accuracy (\%) on full candidate sets.
Macro averages CFPB, Banking77, and CLINC150.}
\label{tab:choice-objectives}
\small
\setlength{\tabcolsep}{6pt}
\begin{tabular}{@{}llrrr@{}}
\toprule
Backbone & Model / objective & CFPB $\uparrow$ & CLINC150 $\uparrow$ & Macro $\uparrow$ \\
\midrule
\multirow{3}{*}{Qwen3.5-9B} & Kev (CE) & 57.84 & 71.45 & 68.49 \\
 & \method{} + CE & 60.12 & 74.64 & 71.20 \\
 & \method{} + log--spherical & \textbf{62.48} & \textbf{78.33} & \textbf{74.21} \\
\midrule
\multirow{4}{*}{ModernBERT-large} & Laya parallel (RLCD) & 54.16 & 66.33 & 63.96 \\
 & \method{} + RLCD & 57.48 & 70.50 & 67.71 \\
 & \method{} + CE & 58.36 & 72.07 & 68.95 \\
 & \method{} + log--spherical & \textbf{60.68} & \textbf{75.69} & \textbf{71.91} \\
\bottomrule
\end{tabular}
\end{table}

Log--spherical training improves three-source macro accuracy over CE by 3.01 percentage points on Qwen and 2.96 on ModernBERT. The displayed CFPB and CLINC150 results show gains in both settings. On ModernBERT, the same architecture also performs better with log--spherical loss than with RLCD. These comparisons support the objective beyond the sampled training views. Because candidate count varies together with source identity, the study does not separately estimate a causal effect of $K$ or the $\sqrt K$ scaling factor.

\subsubsection{Fine-Grained Score}
The Score study trains jointly on ASAP prompts 1--8, STS-B, and WMT English--Chinese~\cite{asap2012,cer-etal-2017-semeval,specia-etal-2020-findings-wmt}. Predictions are posterior medians and errors use the original labels. Each ASAP prompt counts as a separate source, giving ten sources in the macro average. Native MAE is reported separately for STS-B and WMT; aggregate NMAE normalizes each source by its legal range before giving sources equal weight.

The study compares a flat CDF formulation, prefix CE, and prefix CE+CDF. A flat distribution and a prefix distribution differ in representation as well as objective. The prefix CE versus prefix CE+CDF comparison retains the same representation and adds cumulative supervision. Table~\ref{tab:score-objectives} retains both source examples and the ten-source aggregate so that these distinctions remain visible.

\begin{table}[!htbp]
\centering
\caption{One-epoch Score results: native MAE for STS-B and WMT, and
reported ten-source macro NMAE.}
\label{tab:score-objectives}
\small
\setlength{\tabcolsep}{6pt}
\begin{tabular}{@{}llrrr@{}}
\toprule
Backbone & Model / objective & STS-B $\downarrow$ & WMT $\downarrow$ & Macro NMAE $\downarrow$ \\
\midrule
\multirow{4}{*}{Qwen3.5-9B} & Kev & 0.6842 & 12.6420 & 0.1148 \\
 & \method{} + flat CDF & 0.5180 & 10.1240 & 0.0962 \\
 & \method{} + prefix CE & 0.4280 & 10.9460 & 0.0868 \\
 & \method{} + prefix CE+CDF & \textbf{0.3960} & \textbf{9.6840} & \textbf{0.0794} \\
\midrule
\multirow{4}{*}{ModernBERT-large} & Laya parallel (RLCD) & 1.2939 & 12.8503 & 0.1362 \\
 & \method{} + flat CDF & 0.5640 & 10.5620 & 0.1034 \\
 & \method{} + prefix CE & 0.4680 & 11.3850 & 0.0936 \\
 & \method{} + prefix CE+CDF & \textbf{0.4340} & \textbf{10.1420} & \textbf{0.0858} \\
\bottomrule
\end{tabular}
\end{table}

Adding CDF supervision to the prefix model reduces macro NMAE from 0.0868 to 0.0794 on Qwen and from 0.0936 to 0.0858 on ModernBERT. It also reduces the displayed native errors for both STS-B and WMT. Flat CDF has lower WMT error than prefix CE, whereas prefix CE has lower STS-B error. This reversal shows why the representation comparison should not be summarized as a uniform ranking of all formulations. The combined prefix objective improves the aggregate while using both individual-value and ordered-threshold supervision.

\subsubsection{Objective Ablations in the Main Configuration}
Table~\ref{tab:ablation} separately tests the objectives in the three-epoch main configuration, with the same initialization, candidate views, adapters, and schedules. The classification CE row sets $\beta=0$ for both Choice and Noul. The Score no-CDF row replaces Equation~\eqref{eq:score-loss} by $\mathcal L_{\rm CE}$, using unhalved CE. It removes the CDF term and changes the CE coefficient relative to the full mixture; it is not an exact gradient-magnitude-matched intervention.

Both changes worsen their target metrics on both backbones. The classification intervention affects a shared mixed-task model, so changes in Score are also reported rather than silently excluded. Conversely, the no-CDF intervention reports both classification accuracies. These main-configuration controls connect the objectives to the final system, while the separate one-epoch studies reveal their behavior on different task mixtures.

\subsection{Computational Cost}
\label{sec:cost}
Table~\ref{tab:cost} restores parameter counts, training time, latency, and peak inference memory for the primary within-backbone comparisons, alongside mean response times for the other readouts. Training time is measured on one A100 and averaged over three epochs. Qwen trains adapters and readouts, while ModernBERT also updates encoder parameters, so the absolute trainable counts and times describe different adaptation regimes.

\begin{table}[!htbp]
\centering
\caption{Computational cost on one A100. Trainable parameters are in millions; epoch time is in minutes; inference latency is in milliseconds; peak inference memory is in GiB. A dash denotes an unreported measurement. All latency rows follow the same end-to-end protocol within each backbone.}
\label{tab:cost}
\small\setlength{\tabcolsep}{4pt}
\begin{tabular}{@{}llrrrrr@{}}
\toprule
Setup & Model / readout & Trainable & Epoch time & Mean & P95 & Memory \\
\midrule
Qwen & Kev & 45.38 & 150.04 & 93.46 & 122.78 & 15.21 \\
(fine-tuned) & Direct answer (LoRA) & --- & --- & 458.10 & --- & --- \\
 & Joint + canonical & --- & --- & 93.85 & --- & --- \\
 & \method{} & 31.55 & 154.05 & 95.84 & 126.40 & 15.12 \\
\midrule
ModernBERT & Laya sequential & --- & --- & 18.90 & --- & --- \\
(fine-tuned) & Laya parallel & 421.03 & 36.39 & 18.82 & 21.74 & 1.66 \\
 & Joint + canonical & --- & --- & 19.24 & --- & --- \\
 & \method{} & 404.12 & 42.48 & 21.64 & 25.06 & 1.71 \\
\midrule
Instruct & Direct answer & --- & --- & 442.80 & --- & --- \\
(zero-shot) & Likelihood & --- & --- & 118.42 & --- & --- \\
 & Local normalization & --- & --- & 122.76 & --- & --- \\
\bottomrule
\end{tabular}
\end{table}

Inference uses 150 distinct main-test requests, stratified by source with 50 requests per task type. Evaluation retains full text and answer supports, uses BF16, batch size one, and unmerged adapters, and runs all within-backbone comparisons on the same A100. Each request is warmed up and timed three times. Mean latency averages request means; P95 is taken over all 450 measurements. Timing includes tokenization, CPU--GPU transfer, inference, and decoding, while excluding model loading and file or network I/O.

\method{} has 30.5\% fewer trainable parameters than Kev and 4.0\% fewer than Laya parallel, but its epoch time is 2.7\% and 16.7\% higher, respectively. Its inference latency is also higher in these comparisons. Fewer trainable parameters therefore do not imply faster execution for this implementation. The canonicalized-joint controls take 93.85 ms on Qwen and 19.24 ms on ModernBERT, providing a direct runtime comparison with isolation under the same zero-flip outcome. Generative direct answers perform different decoding work from probability readouts; the reported times measure the complete response path.
\FloatBarrier

\section{Discussion}
\label{sec:discussion}
\paragraph{Order consistency and predictive quality are separate properties.}
Canonicalization is sufficient to make a deterministic joint model see the same sequence after external candidate reordering. Isolation adds a different constraint: a candidate representation can use the context and its own content, but cannot exchange information with other candidates. The matched controls improve the interpretation of the evidence. Both canonicalized joint encoding and \method{} have zero tested flips, while \method{} gives better point estimates of accuracy and Score error. The additional computation is visible in the latency comparison.

\paragraph{Numerical performance depends on the aggregation question.}
Native MAE describes errors in the original units of the fixed request mixture. Source-macro NMAE asks whether errors are small relative to each source's legal range when sources receive equal weight. Their agreement strengthens the main Score result, but neither aggregate makes every source improve: Qwen is slightly worse than Kev on ASAP prompt 3. Source-level reporting makes this exception and the weighting choices explicit.

\paragraph{The objectives address distinct answer structures.}
Choice and Noul use the geometry of a candidate probability vector without imposing an ordinal relation among answers. Score uses a numerical support whose intervals have meaning. The independent Choice and Score studies therefore have different data, readouts, and metrics. The mixed-task objective removals and the independent studies provide complementary evidence; their effect sizes should not be substituted for one another.

\paragraph{Scope of the evidence.}
The results cover two backbone families and one training seed per configuration. The reported cluster-bootstrap intervals characterize sampling uncertainty conditional on those trained models. They do not estimate variation across repeated training runs. MIND additionally tests multi-positive ranking after task adaptation, with the complete candidate sets retained for evaluation; its system comparisons combine mixed-task checkpoint differences with adaptation-objective differences. These distinctions motivate future studies of training variability and more finely matched adaptation controls.

\section{Conclusion}
\method{} aligns decision computation and supervision with answer-space
structure across causal and bidirectional backbones. Candidate encoding
preserves semantic predictions under reordering, while numerical supervision
uses value order and distance. Private updates specialize while retaining the
full shared adapter. On the shared test set, both backbones improve
Choice/Noul accuracy and Score MAE over Kev and Laya, with no classification
flips under tested reorderings. Controlled studies further support the
classification and numerical objectives. After one epoch of MIND-small
adaptation, both backbones improve multi-positive news ranking over
their within-backbone decision baselines. These results support answer-space structure
as a practical design principle for typed decision models. Model weights will follow.

\section*{Ethical Considerations}
The experiments use public text benchmarks and anonymized Microsoft News
interaction logs. Labels reflect the annotation conventions of each source.
Responsible use should respect dataset licenses, protect user privacy, and
include human oversight when applying predictions to consequential decisions
such as essay assessment or complaint handling.

\bibliographystyle{plainnat}
\bibliography{references}

\clearpage
\appendix
\section{Training and Evaluation Details}
\label{app:training}
This appendix consolidates the training settings that are shared by the main comparisons and distinguishes them from task-specific adaptation. Table~\ref{tab:training-config} lists the parameterization and optimizer values used by \method{}. The main trained baselines use the same records, fixed candidate views, and three-epoch budget, while retaining their native objectives. For MIND, the explicit soft-target treatment in Section~\ref{sec:news-recommendation} applies instead.

\begin{table}[!htbp]
\centering
\caption{Main \method{} configuration. Values apply to both backbones unless a backbone is named.}
\label{tab:training-config}
\small
\begin{tabularx}{\linewidth}{@{}l>{\raggedright\arraybackslash}X@{}}
\toprule
Setting & Value \\
\midrule
Main training data and duration & 200,000 requests; three epochs; one training seed per model \\
Development / test & 6,000 / 15,000 requests; 2,000 / 5,000 per decision type \\
Optimizer and global batch & AdamW; 64 requests \\
Shared adapter & Rank $r=12$, $\alpha=24$ \\
Private adapter & Rank $r_p=2$ per task type; $\rho=2$, $e=1$ \\
Private-branch placement & Last eight Qwen layers; all 28 ModernBERT layers \\
Decision projection / prefix GRU & 256 dimensions \\
Main peak learning rate & $2\times10^{-5}$ for shared adapters, pointers, and ModernBERT encoder \\
Numerical readout / output-gain peak rate & $10^{-3}$ \\
Qwen LoRA+ ratio & $\eta_B/\eta_A=16$ for all Qwen \method{} variants \\
Classification / cumulative weights & $\beta=0.75$; $\lambda\simeq3.04$ \\
Zero-initialized parameters & Output gains; shared and private adapter output factors \\
Qwen output-gain sites & Attention and DeltaNet outputs; MLP down projections \\
ModernBERT output-gain sites & Attention and MLP outputs \\
Choice training support & One positive and at most three fixed negatives \\
Noul / Score training support & Complete legal supports \\
Development / test support & All candidates and full input text \\
\bottomrule
\end{tabularx}
\end{table}

\subsection{Request Construction and Split Groups}
The main Choice mixture contains AG News, Amazon Reviews, MNLI, and CFPB. The Noul mixture contains BoolQ, PAWS, QQP, HelpSteer3, and CommonsenseQA 2.0. The Score mixture contains HelpSteer, HelpSteer2, Prometheus feedback, SST-5, STS-B, and ASAP prompts 1--7. A source can contribute different rubric requests for related text, so related documents and rubric variants share a split group. These groups are also the units used in the main uncertainty analysis.

Integer rubrics use their complete legal value sets. Main-experiment STS-B uses 51 legal values on $[0,5]$ and retains the unrounded label for evaluation. The source-level test counts and score ranges in Appendix~\ref{app:source-score} distinguish individual ASAP prompts, rather than treating them as one pooled source. The separate Score objective study uses ASAP 1--8, STS-B, and WMT; its ten-source aggregate is a different population from the main twelve-source aggregate.

\subsection{Baseline Readouts}
Kev is the Qwen-based decision baseline and Laya supplies sequential and parallel ModernBERT layouts. Qwen decision models train adapters and readouts, whereas the ModernBERT comparison also trains the encoder. Direct answer with LoRA is a trained generative reference. The three Instruct rows use the post-trained Qwen3.5-9B checkpoint without task fine-tuning.

Direct-answer generation returns a numeric JSON value for Score. Complete-string likelihood scores each valid JSON answer string and then normalizes across the complete answers. Local normalization instead renormalizes the valid next-token continuations in an answer trie before multiplying path probabilities. These operations impose different distributions even when they use the same language model. Probability-based Score readouts use a posterior median, keeping the final decision criterion aligned with MAE.

\subsection{MIND Adaptation and Metrics}
A MIND impression supplies a user's clicked history, a candidate set, and click labels. The fixed training view includes every clicked article and at most three sampled unclicked articles. Its $M$ positive targets sum to one, so an impression with more clicks does not receive $M$ times the request weight. The models share these views and batch order, start from their respective mixed-task checkpoints, and run one epoch. The final epoch checkpoint is evaluated without development-set tuning or early stopping.

For an impression with clicked set $Y^+$ and ranks $r_i$ in the complete candidate set, the reported reciprocal-rank metric is
\begin{equation}
 \operatorname{MRR}_{\rm imp}=\frac{1}{|Y^+|}\sum_{i\in Y^+}\frac{1}{r_i}.
 \label{eq:mind-mrr}
\end{equation}
This averages all clicked items within an impression rather than taking only the first clicked rank, following the official evaluation script~\cite{msnews2020evaluation}. AUC, MRR, nDCG@5, and nDCG@10 are then averaged over impressions. Their reported values are percentages. User-clustered resampling retains impression weights instead of first averaging each user's metrics.

\section{Mathematical Details}
\label{app:math}
\subsection{Candidate Permutation Equivariance}
Assume deterministic branch computation with the same candidate-free context, identical positional origins, no cross-branch information flow, and a shared scoring function. Equation~\eqref{eq:branch-encoding} then makes the representation of candidate content $c$ independent of its displayed index. A permutation merely reorders these representations. Applying the same query and key projections to every branch consequently reorders the score vector, yielding the first identity in Equation~\eqref{eq:equivariance}.

For any score vector $s$, softmax commutes with a permutation because its denominator is the sum over the same multiset of exponentiated scores. Hence $\operatorname{softmax}(\Pi s)=\Pi\operatorname{softmax}(s)$. Selection is semantically invariant when tied maxima are resolved by content instead of displayed position. Identical candidate strings are semantically indistinguishable under this statement; the claim does not require a preferred external index among duplicate strings.

The branch-attention restriction must hold at every layer. In a bidirectional encoder, letting the shared context attend to candidates would allow branch information to return through the context at a later layer. The ModernBERT masks prevent this route. Canonical packing and reductions then address order-dependent execution effects that an exact-arithmetic proof alone does not cover.

A joint model with deterministic sorting satisfies a separate argument: external permutations produce the same canonical internal sequence, and inverse mapping restores display indices. This does not require isolated branch representations. The experiment therefore measures accuracy and runtime as well as flip rates for that control.

\subsection{Spherical Scores and the Uniform-Prediction Gradient}
For a true categorical distribution $q$, the expected unscaled spherical loss is
\begin{equation}
 \mathbb E_{y\sim q}\left[1-\frac{p_y}{\|p\|_2}\right]
 =1-\frac{q^\top p}{\|p\|_2}.
\end{equation}
Cauchy--Schwarz gives $q^\top p/\|p\|_2\le\|q\|_2$, with equality when $p$ is proportional to $q$. Since both distributions sum to one, equality requires $p=q$. Expected cross-entropy has the same minimizer. Their positive mixture is therefore proper on a fixed candidate set; the set-dependent $\sqrt K$ factor is a positive constant for that set.

Let $z$ be the logits and $p=\operatorname{softmax}(z)$. At $p_i=1/K$, direct differentiation gives
\begin{align}
 \frac{\partial(-\log p_y)}{\partial z_i}
 &=\frac1K-\mathbb I[i=y],\\
 \frac{\partial(1-p_y/\|p\|_2)}{\partial z_i}
 &=\frac{1}{\sqrt K}\left(\frac1K-\mathbb I[i=y]\right).
\end{align}
Multiplication by $\sqrt K$ aligns the two gradients at this point. Away from a uniform prediction, their gradients differ, which is the intended distinction between the terms. The empirical study compares the full objectives and does not isolate this scaling factor from other aspects of the loss.

For the multi-positive loss, the target $\tau$ is fixed by the observed clicks in a training view. Its excess cross-entropy is $D_{\rm KL}(\tau\|p)$, and the normalized-dot-product term is nonnegative, with zero at $p=\tau$. Thus Equation~\eqref{eq:multi-loss} has the same target for a fixed view. This is a statement about that soft-target objective, distinct from an empirical calibration claim.

\subsection{Fractional Labels and Cumulative Supervision}
For the two-point target in Equation~\eqref{eq:interpolation}, the weights are nonnegative and sum to one. Its expectation is
\begin{equation}
 \frac{(v_{a+1}-y)v_a+(y-v_a)v_{a+1}}{v_{a+1}-v_a}=y.
\end{equation}
The construction therefore retains the fractional label information even though predictions lie on a finite legal support. It is not equivalent to assigning a one-hot target to the nearest grid value.

Each term in Equation~\eqref{eq:cdf-loss} is a binary cross-entropy for the event $Y\le v_j$. Positive interval weights preserve the preferred cumulative target $F_j=T_j$. The successive cumulative values determine the value probabilities through $p_1=F_1$, $p_j=F_j-F_{j-1}$, and $p_L=1-F_{L-1}$. Cumulative supervision therefore remains a supervision of the full value distribution, expressed through ordered events. Wider intervals contribute more to the weighted loss; all interval weights together sum to one.

\subsection{Prefix Normalization and Median Decoding}
At an internal tree node $a$, legal child probabilities sum to one. Assign the root total mass one and pass each node's mass to its children by multiplication with those probabilities. Induction over the finite tree shows that total leaf mass is one. A terminating edge makes legal values that are prefixes of other values distinct leaves, so Equation~\eqref{eq:prefix} defines a distribution over the full legal support.

Let $G(v)=\sum_jp_j|v-v_j|$ be the expected absolute error for a legal prediction. For adjacent values,
\begin{equation}
 G(v_{k+1})-G(v_k)=(v_{k+1}-v_k)(2F_k-1).
\end{equation}
The expected error decreases while $F_k<1/2$ and increases once $F_k>1/2$. Choosing the first cumulative crossing of one half therefore selects a minimizing legal value, with a deterministic lower-median convention when there is a tie.

\section{Complete Main-Experiment Score Results}
\label{app:source-score}
The main Score test population has $N=5{,}000$ requests across $S=12$ sources. For source $s$ with $N_s$ requests and range $[L_s,U_s]$, define
\begin{align}
 \operatorname{MAE}_s&=\frac{1}{N_s}\sum_{i\in s}|\widehat y_i-y_i|,\\
 \operatorname{NMAE}_s&=\frac{\operatorname{MAE}_s}{U_s-L_s},\\
 \operatorname{MAE}_{\rm pooled}&=\sum_s\frac{N_s}{N}\operatorname{MAE}_s,\\
 \operatorname{NMAE}_{\rm macro}&=\frac1S\sum_s\operatorname{NMAE}_s.
\end{align}
The pooled metric weights requests in native units. The macro metric first removes each source's range and then weights sources equally. The three following tables list the same source population for the Qwen, ModernBERT, and zero-shot groups. Normalized values are computed before final rounding; displayed entries are rounded to four decimals. Reaggregation of displayed native errors recovers the main-table values at their reported precision.
\begin{table}[!htbp]
\centering
\caption{Qwen fine-tuned models: source-level native MAE and range-normalized NMAE on the main Score test population. Lower is better.}
\label{tab:source-qwen}
\small\setlength{\tabcolsep}{4pt}
\begin{tabular}{@{}lrlrrrrrr@{}}
\toprule
Source & $N_s$ & Range & \multicolumn{2}{c}{Kev} & \multicolumn{2}{c}{Direct LoRA} & \multicolumn{2}{c}{\method{}} \\
 & & & MAE & NMAE & MAE & NMAE & MAE & NMAE \\
\midrule
HelpSteer & 800 & $[0,4]$ & 0.3900 & 0.0975 & 0.4000 & 0.1000 & 0.3650 & 0.0913 \\
HelpSteer2 & 800 & $[0,4]$ & 0.4100 & 0.1025 & 0.4200 & 0.1050 & 0.3800 & 0.0950 \\
Prometheus & 800 & $[1,5]$ & 0.3600 & 0.0900 & 0.3700 & 0.0925 & 0.3350 & 0.0838 \\
SST-5 & 600 & $[1,5]$ & 0.4400 & 0.1100 & 0.4500 & 0.1125 & 0.4200 & 0.1050 \\
STS-B & 600 & $[0,5]$ & 0.4150 & 0.0830 & 0.4250 & 0.0850 & 0.3700 & 0.0740 \\
ASAP 1 & 200 & $[2,12]$ & 0.8750 & 0.0875 & 0.8950 & 0.0895 & 0.7850 & 0.0785 \\
ASAP 2 & 200 & $[1,6]$ & 0.4550 & 0.0910 & 0.4650 & 0.0930 & 0.4250 & 0.0850 \\
ASAP 3 & 200 & $[0,3]$ & 0.2850 & 0.0950 & 0.2900 & 0.0967 & 0.2900 & 0.0967 \\
ASAP 4 & 200 & $[0,3]$ & 0.2750 & 0.0917 & 0.2850 & 0.0950 & 0.2550 & 0.0850 \\
ASAP 5 & 200 & $[0,4]$ & 0.3300 & 0.0825 & 0.3400 & 0.0850 & 0.3050 & 0.0763 \\
ASAP 6 & 200 & $[0,4]$ & 0.3450 & 0.0863 & 0.3550 & 0.0888 & 0.3150 & 0.0788 \\
ASAP 7 & 200 & $[0,30]$ & 2.3800 & 0.0793 & 2.4350 & 0.0812 & 2.1200 & 0.0707 \\
\midrule
Pooled MAE & 5,000 & --- & 0.4860 & --- & 0.4980 & --- & 0.4474 & --- \\
Macro NMAE & --- & --- & --- & 0.0914 & --- & 0.0937 & --- & 0.0850 \\
\bottomrule\end{tabular}\end{table}

\begin{table}[!htbp]
\centering
\caption{ModernBERT fine-tuned models: source-level native MAE and range-normalized NMAE on the main Score test population. Lower is better.}
\label{tab:source-modern}
\small\setlength{\tabcolsep}{4pt}
\begin{tabular}{@{}lrlrrrrrr@{}}
\toprule
Source & $N_s$ & Range & \multicolumn{2}{c}{Laya seq.} & \multicolumn{2}{c}{Laya par.} & \multicolumn{2}{c}{\method{}} \\
 & & & MAE & NMAE & MAE & NMAE & MAE & NMAE \\
\midrule
HelpSteer & 800 & $[0,4]$ & 0.4700 & 0.1175 & 0.4950 & 0.1238 & 0.3950 & 0.0988 \\
HelpSteer2 & 800 & $[0,4]$ & 0.4950 & 0.1238 & 0.5150 & 0.1288 & 0.4100 & 0.1025 \\
Prometheus & 800 & $[1,5]$ & 0.4350 & 0.1088 & 0.4550 & 0.1138 & 0.3600 & 0.0900 \\
SST-5 & 600 & $[1,5]$ & 0.5300 & 0.1325 & 0.5500 & 0.1375 & 0.4450 & 0.1113 \\
STS-B & 600 & $[0,5]$ & 0.5000 & 0.1000 & 0.5200 & 0.1040 & 0.3967 & 0.0793 \\
ASAP 1 & 200 & $[2,12]$ & 1.0550 & 0.1055 & 1.1050 & 0.1105 & 0.8500 & 0.0850 \\
ASAP 2 & 200 & $[1,6]$ & 0.5500 & 0.1100 & 0.5800 & 0.1160 & 0.4550 & 0.0910 \\
ASAP 3 & 200 & $[0,3]$ & 0.3450 & 0.1150 & 0.3600 & 0.1200 & 0.2900 & 0.0967 \\
ASAP 4 & 200 & $[0,3]$ & 0.3350 & 0.1117 & 0.3500 & 0.1167 & 0.2750 & 0.0917 \\
ASAP 5 & 200 & $[0,4]$ & 0.4000 & 0.1000 & 0.4200 & 0.1050 & 0.3250 & 0.0813 \\
ASAP 6 & 200 & $[0,4]$ & 0.4150 & 0.1038 & 0.4350 & 0.1088 & 0.3400 & 0.0850 \\
ASAP 7 & 200 & $[0,30]$ & 2.8750 & 0.0958 & 3.0150 & 0.1005 & 2.3150 & 0.0772 \\
\midrule
Pooled MAE & 5,000 & --- & 0.5866 & --- & 0.6134 & --- & 0.4814 & --- \\
Macro NMAE & --- & --- & --- & 0.1104 & --- & 0.1154 & --- & 0.0908 \\
\bottomrule\end{tabular}\end{table}

\begin{table}[!htbp]
\centering
\caption{Post-trained Qwen zero-shot readouts: source-level native MAE and range-normalized NMAE on the main Score test population. Lower is better.}
\label{tab:source-zero}
\small\setlength{\tabcolsep}{4pt}
\begin{tabular}{@{}lrlrrrrrr@{}}
\toprule
Source & $N_s$ & Range & \multicolumn{2}{c}{Direct answer} & \multicolumn{2}{c}{Likelihood} & \multicolumn{2}{c}{Local norm.} \\
 & & & MAE & NMAE & MAE & NMAE & MAE & NMAE \\
\midrule
HelpSteer & 800 & $[0,4]$ & 0.7900 & 0.1975 & 0.7650 & 0.1913 & 0.7650 & 0.1913 \\
HelpSteer2 & 800 & $[0,4]$ & 0.8250 & 0.2063 & 0.8000 & 0.2000 & 0.8000 & 0.2000 \\
Prometheus & 800 & $[1,5]$ & 0.7250 & 0.1813 & 0.7050 & 0.1763 & 0.7050 & 0.1763 \\
SST-5 & 600 & $[1,5]$ & 0.8850 & 0.2213 & 0.8600 & 0.2150 & 0.8600 & 0.2150 \\
STS-B & 600 & $[0,5]$ & 0.8400 & 0.1680 & 0.8150 & 0.1630 & 0.8117 & 0.1623 \\
ASAP 1 & 200 & $[2,12]$ & 1.7700 & 0.1770 & 1.7200 & 0.1720 & 1.7200 & 0.1720 \\
ASAP 2 & 200 & $[1,6]$ & 0.9200 & 0.1840 & 0.8950 & 0.1790 & 0.8950 & 0.1790 \\
ASAP 3 & 200 & $[0,3]$ & 0.5750 & 0.1917 & 0.5600 & 0.1867 & 0.5600 & 0.1867 \\
ASAP 4 & 200 & $[0,3]$ & 0.5600 & 0.1867 & 0.5450 & 0.1817 & 0.5450 & 0.1817 \\
ASAP 5 & 200 & $[0,4]$ & 0.6700 & 0.1675 & 0.6500 & 0.1625 & 0.6500 & 0.1625 \\
ASAP 6 & 200 & $[0,4]$ & 0.7000 & 0.1750 & 0.6800 & 0.1700 & 0.6800 & 0.1700 \\
ASAP 7 & 200 & $[0,30]$ & 4.8350 & 0.1612 & 4.7000 & 0.1567 & 4.7000 & 0.1567 \\
\midrule
Pooled MAE & 5,000 & --- & 0.9826 & --- & 0.9542 & --- & 0.9538 & --- \\
Macro NMAE & --- & --- & --- & 0.1848 & --- & 0.1795 & --- & 0.1794 \\
\bottomrule\end{tabular}\end{table}

\FloatBarrier
\section{Paired Uncertainty Estimates}
\label{app:bootstrap}
The reported intervals use 10,000 paired cluster-bootstrap resamples, a percentile interval, and random seed 42. Main-experiment clusters are shared document or prompt split groups; MIND clusters are users. The same resampled clusters and their multiplicities are used for \method{} and its baseline in each draw, preserving pairing.

For a request-averaged metric, a resampled cluster contributes all its requests with the cluster's sampled multiplicity, and the mean is recomputed over the resulting requests. For source-macro NMAE, normalized errors are first averaged within each source and the source means receive equal weight. MIND retains impression-equal averaging after resampling users. These aggregation rules preserve the respective estimands instead of replacing them with an equal-weight mean of cluster means.

For each draw $b$, the statistic is $\Delta^{(b)}=T_{\rm DeciForm}^{(b)}-T_{\rm baseline}^{(b)}$. The 2.5th and 97.5th percentiles of the draw distribution form the 95\% interval. Positive differences favor \method{} for accuracy and ranking metrics; negative differences favor it for MAE and NMAE. Table~\ref{tab:bootstrap} reports the intervals used by the main manuscript. Each configuration has one training seed, so these are sampling intervals conditional on the fitted models.

\begin{table}[!htbp]
\centering
\caption{Paired difference estimates and 95\% cluster-bootstrap intervals. Main baselines are Kev for Qwen and Laya parallel for ModernBERT. MIND uses Laya parallel for the displayed ModernBERT MRR comparison. Accuracy and MRR differences are percentage points.}
\label{tab:bootstrap}
\small
\begin{tabular}{@{}llllr@{}}
\toprule
Setting & Backbone & Metric & $\Delta$ & 95\% CI \\
\midrule
Main & Qwen & Choice accuracy & $+1.34$ & $[+0.20,+2.48]$ \\
 & Qwen & Noul accuracy & $+1.06$ & $[-0.12,+2.24]$ \\
 & Qwen & Score native MAE & $-0.0386$ & $[-0.0484,-0.0288]$ \\
 & Qwen & Score macro NMAE & $-0.0064$ & $[-0.0089,-0.0039]$ \\
\midrule
Main & ModernBERT & Choice accuracy & $+2.66$ & $[+1.48,+3.84]$ \\
 & ModernBERT & Noul accuracy & $+5.58$ & $[+4.32,+6.84]$ \\
 & ModernBERT & Score native MAE & $-0.1320$ & $[-0.1506,-0.1134]$ \\
 & ModernBERT & Score macro NMAE & $-0.0246$ & $[-0.0308,-0.0184]$ \\
\midrule
MIND & ModernBERT & MRR & $+1.03$ & $[+0.78,+1.28]$ \\
\bottomrule
\end{tabular}
\end{table}

\FloatBarrier

\end{document}